\documentclass[runningheads]{llncs}
\usepackage[T1]{fontenc}
\usepackage{graphicx}
\usepackage{multirow}
\usepackage{makecell}
\usepackage{dsfont}
\usepackage{amssymb}
\begin{document}
\title{Revisiting Human-in-the-Loop Object Retrieval with Pre-Trained Vision Transformers}
\titlerunning{Revisiting Human-in-the-Loop Object Retrieval with Pre-Trained ViTs}
%
\author{Kawtar Zaher\inst{1, 2} \and
Olivier Buisson\inst{1} \and
Alexis Joly\inst{2}}
\authorrunning{K. Zaher et al.}
%
\institute{Institut National de l'Audiovisuel INA, \\ Paris, France \and
Institut National de Recherche et Informatique et Robotique INRIA, \\ Montpellier, France}
\maketitle              
\begin{abstract}
Building on existing approaches, we revisit Human-in-the-Loop Object Retrieval, a task that consists of iteratively retrieving images containing objects of a class-of-interest, specified by a user-provided query. Starting from a large unlabeled image collection, the aim is to rapidly identify diverse instances of an object category relying solely on the initial query and the user's Relevance Feedback, with no prior labels. The retrieval process is formulated as a binary classification task, where the system continuously learns to distinguish between relevant and non-relevant images to the query, through iterative user interaction. This interaction is guided by an Active Learning loop: at each iteration, the system selects informative samples for user annotation, thereby refining the retrieval performance. This task is particularly challenging in multi-object datasets, where the object of interest may occupy only a small region of the image within a complex, cluttered scene. Unlike object-centered settings where global descriptors often suffice, multi-object images require more adapted, localized descriptors. In this work, we formulate and revisit the Human-in-the-Loop Object Retrieval task by leveraging pre-trained ViT representations, and addressing key design questions, including which object instances to consider in an image, what form the annotations should take, how Active Selection should be applied, and which representation strategies best capture the object's features. We compare several representation strategies across multi-object datasets highlighting trade-offs between capturing the global context and focusing on fine-grained local object details. Our results offer practical insights for the design of effective interactive retrieval pipelines based on Active Learning for object class retrieval.
 
\end{abstract}
\section{Introduction}

Object Retrieval (OR)~\cite{deselaers2008overview} aims to identify and retrieve images containing objects from predefined categories. Traditional OR relies on a single retrieval step, which often falls short, especially in contexts where the desired concept in not very common. This is typically the case of specialized and underrepresented expert domains, like audio-visual archives, digitized historical collections, etc. For example, a historian interested in car images from a specific era might initially get results showing general car models. Iterative refinement becomes crucial to progressively narrow down the search to the specific era of interest. 
\vspace{-1.5em}
\begin{figure}[h]
    \centering
    \includegraphics[width=0.7\linewidth]{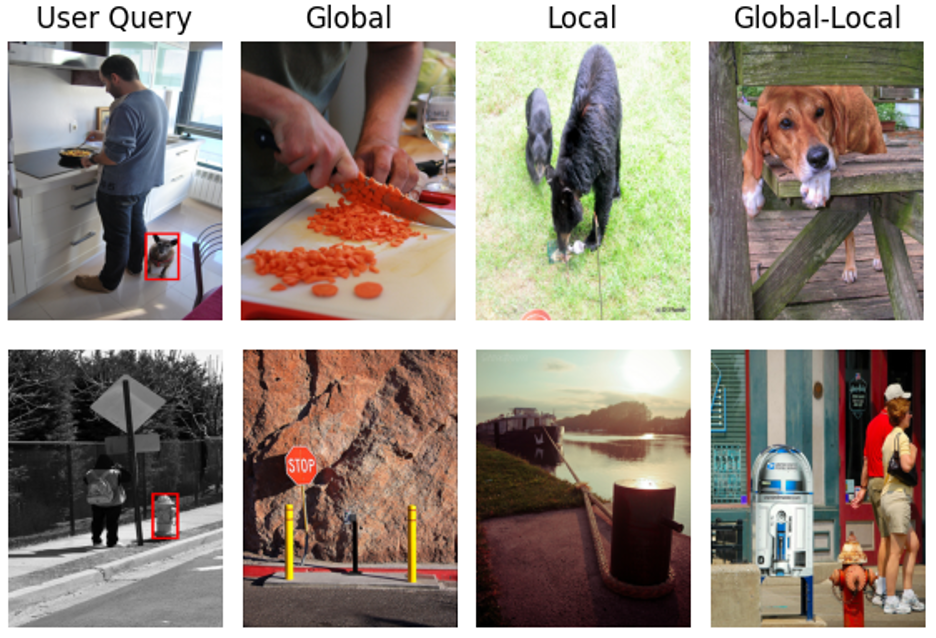}
    \caption{\centering Retrieved images during earlier iterations for two initial queries ("dog", "fire hydrant"): the global descriptor captures the dominating overall concept("cooking", "road"), the local one recognizes the parts only ("furry animal", "cylindric object"), whereas the hybrid one allows to capture the desired object.}
    \label{fig:example}
\end{figure}
\vspace{-1.5em} \\
\noindent Human-in-the-Loop OR addresses this issue by incorporating a Human-in-the-Loop's~\cite{mosqueira2023human} feedback about the relevant regions of interest. To our knowledge, not many works addressed this setup. The work of~\cite{litayem2009interactive} introduces an active learning-based framework for interactive object retrieval, where the system actively selects the most informative samples for labeling by the user. The ability to incorporate the user's feedback allows the retrieval process to be refined progressively, relying only on a small number of labeled samples. However, this framework only relies on handcrafted features, and adequate SIFT matching techniques. Deep learning techniques on the other hand, present great potential to be leveraged in this context.  

\begin{figure}[h]
    \centering
    \includegraphics[width=1\linewidth]{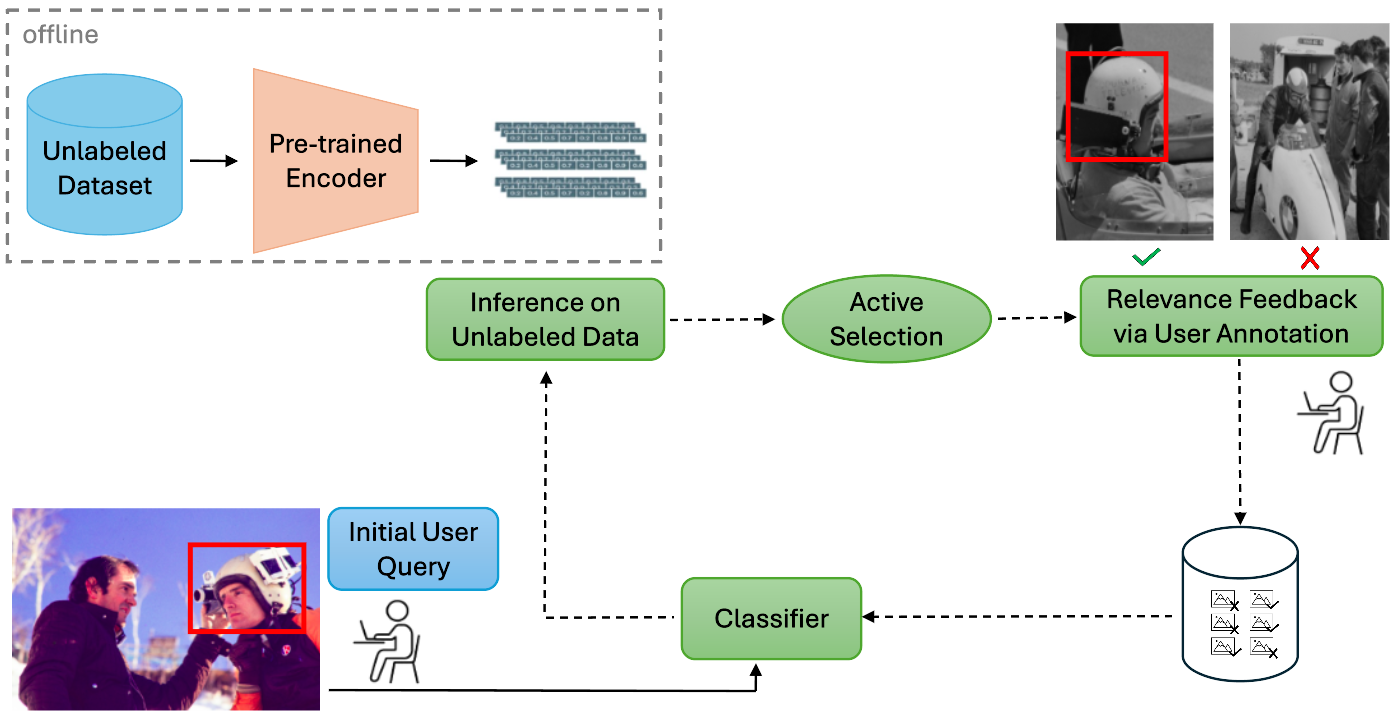}
    \caption{\centering Visual representation of the Human-in-the-Loop Object Retrieval framework.}
    \label{schema}
\end{figure}

Building on this idea, we revisit the Human-in-the-Loop Object Retrieval task (HITL-OR), presented in Fig.~\ref{schema}, which integrates an Active Learning (AL) loop to guide the retrieval process. HITL-OR allows a user to interactively explore large unlabeled datasets through an iterative retrieval process. This process identifies and retrieves images containing objects that conform to the pattern of an initial user-provided query, labeled with a region of interest, without requiring prior training or prior labels. Offline, the process begins with data preparation, where pre-trained representations are extracted to describe the dataset images. Online, a shallow binary classifier is trained to distinguish the class-of-interest with few shots labeled iteratively. At each iteration, the trained classifier scores the unlabeled data, and informative samples are selected using an AL criterion. The user then provides feedback on the retrieved images, identifying relevant samples and selecting a region of interest in each of them. This feedback updates the classifier, and the process repeats, progressively improving retrieval performance until the user is satisfied or a maximum number of iterations is reached.
\vspace{-1.5em}
\vspace{-1.5em} \\
\noindent To make the most of the limited feedback provided by the user, the retrieval stage leverages Active Learning (AL) strategies. Since the user can label only a small number of samples per iteration, selecting the most informative samples is crucial. AL allows the system to prioritize examples that are expected to most improve the classifier's ability to recognize the class-of-interest, thereby enhancing the retrieval performance under minimal supervision. 

Our contributions can be summarized as follows: 1) We revisit the Human-in-the-Loop Object Retrieval task, addressing key components such as: object selection, user annotation format, Active Learning application, and representation choices. 2) We compare global, local and hybrid representations in the Human-in-the-Loop Object Retrieval setup for multi-object datasets, highlighting the trade-offs between the global context information and the fine-grained details in our setting. 3) We perform different experiments to evaluate our framework, demonstrating the effectiveness of hybrid representations in achieving high retrieval performance with minimal user input.

\section{Related Work}
\subsubsection{Object Retrieval.} 
Object Retrieval (OR) is the task of identifying and returning images containing objects from a category specified by a query. Unlike Object Detection, which localizes all object instances within a single image, or Object Recognition, which classifies an image into a predefined set of classes, retrieval operates at the dataset level, searching for matching instances across a large dataset. OR methods fall into two categories: instance-based retrieval~\cite{arandjelovic2012three,tolias2015particular}, which seeks to find the exact same object instance, and category-level retrieval~\cite{litayem2009interactive,deselaers2008overview}, in which all different and diverse object instances are relevant. Our work focuses on the latter. Early OR relied on handcrafted features like SIFT~\cite{lowe2004distinctive}, with aggregation methods like BoVW~\cite{csurka2004visual}, Fisher Vectors~\cite{perronnin2007fisher}, and VLAD~\cite{jegou2010aggregating} for compact descriptors. Deep learning retrieval works~\cite{chen2022deep} later replaced these descriptors with pre-trained CNN features~\cite{razavian2014cnn} and aggregation methods like R-MAC~\cite{tolias2015particular} and NetVLAD~\cite{arandjelovic2016netvlad} offering richer representations and better performances. More recent works~\cite{el2021training,tan2021instance,song2023boosting} explore ViT-based representations. However, these methods assume prior knowledge of object categories during training on retrieval datasets, while our approach starts with no such knowledge. Moreover, there are many contexts in which a single instance of the object is insufficient because it does not capture the variability of the concept, as mentioned before. 

\subsubsection{Human-in-the-Loop Retrieval and Relevance Feedback using Active Learning.} 
HITL Retrieval methods enhance OR by incorporating human feedback to refine the retrieval. In unannotated datasets, Relevance Feedback (RF)~\cite{patil2011relevance} allows users to continuously specify relevant images, to guide the retrieval process. Early RF works queried the user on top-ranked samples, which provided poor additional knowledge to the classifier. Active Learning (AL)~\cite{settles2009active} was later integrated to the RF framework~\cite{tong2001support,ferecatu2007interactive,demir2014novel,ngo2016image}, prioritizing the most informative samples to reduce the number of required labels for significant improvements. But theses works only performed on whole images, disregarding local Object Retrieval.

\subsubsection{Few-Shot Learning (FSL).} Our Interactive Object Retrieval task bears similarities to FSL~\cite{wang2020generalizing}, due to the binary classification component with scarce labels. Both tasks aim to generalize to an unlabeled dataset from only a few examples. In a $2$-way $k$-shot FSL setting, a model is trained to classify unlabeled queries into 2 classes, using only a limited number $k$ of samples per class. While FSL chooses the samples of the support set at random, our problem is framed within an AL setting, where the samples are selected following a certain criterion. Active Few Shot Learning (AFSL)~\cite{abdali2023active} focuses on selecting the samples of the support set with an Active Selection procedure. The most informative samples to classify the query set within the classes of the support set are selected via an AL criterion, and the FSL learning is then conducted in a traditionnal manner. Hence, AFSL remains classification-based, focusing more on actively selecting samples that improve the classification accuracy for all classes at once. In contrast, our task is more user-centered, prioritizing the samples relevance to one user-provided concept, emphasizing retrieval over classification.

\subsubsection{Multi-Object Image Understanding in the Context of Self-supervised Image Retrieval.} 
Retrieving a specific object from multi-object images with cluttered backgrounds requires selecting an appropriate representation. Global descriptors~\cite{shao2023global}, obtained from CNN~\cite{o2015introduction} or ViTs~\cite{dosovitskiy2020image} last layers, compress the entire image into a single vector, capturing overall scene context efficiently but often including irrelevant background and other objects, overlooking small or occluded objects of interest. Local descriptors, derived from CNN activations or ViT patch tokens, focus on specific regions or patches, retaining fine-grained details but lacking global context, and producing a large number of vectors per image. Object-level descriptors, obtained via segmentation masks~\cite{garg2024revisit} or bounding boxes, provide precise object representations but are limited to predefined classes, making them unsuitable for complex user-defined concepts (e.g., "a person with a red hat walking a dog"). Patch-level descriptors computed on fixed grids offer a compromise, providing systematic coverage without relying on external segmentation or detection models. To further capture both global context and local details, various fusion strategies, targeted for IR, have been proposed, such as DELG~\cite{cao2020unifying} (2-stage IR), DOLG~\cite{yang2021dolg}, DALG~\cite{song2022dalg}, DTop~\cite{song2023boosting}. However, these methods primarily require supervised training and/or fine-tuning on the target dataset, as well as pre-learned fixed fusions specific to certain objects in the image, disregarding the user input. Our fusion strategies are rather online, to take into account the desired concept position within the image. Other relevant self-supervised works like~\cite{govindarajan2021self} specialize pre-trained features to the target dataset. Inspired by recent studies~\cite{denner2025leveraging}, we leverage the high potential of self-supervised foundation models. In our framework, we adopt fixed-grid patch descriptors from foundation ViTs with learning-free, online fusion strategies to ensure both taking into account the user queries and applicability to multi-object, human-in-the-loop retrieval. 

\section{The Human-in-the-Loop Object Retrieval Framework}

\subsection{General Task Definition}

The initial dataset $D$ is a set of $N$ unlabeled images $D = \{I_i\}_{i \in \{1, 2, ..., N\}}$. We initially focus on the case of global descriptors to provide a clear understanding of the framework steps.
\begin{enumerate}
    \setlength{\itemsep}{0em}
    \item Each image $I_i$ is encoded using a pretrained neural network $\Phi$ to obtain a global descriptor $g_i$: \vspace{-0.5em} 
    \begin{equation}
        g_i = \Phi(I_i) \in \mathds{R}^d
    \end{equation}
    \item The user provides an input query that belongs the class-of-interest $c$. The goal is to iteratively retrieve images of $D$ containing objects that belong to the same class $c$.
    \item The provided query initiates a labeled training set $D_l$, and is labeled as 1 (positive class). Under the assumption of a large dataset with many classes, a random number of images is sampled, assumed as outside the class-of-interest, labeled as 0 (negative class) and added to $D_l$.
    \item A shallow linear SVM classifier $f$ is then trained on $D_l$ to discriminate between the class-of-interest and the rest of the dataset. As per~\cite{aggarwal2022optimizing}, we chose an SVM rather than fine-tuning a linear layer using SGD: in settings where the annotation budget is constrained, SVMs are known to better generalize on binary classification problems.
    \item A prediction score $s_i$ is obtained for each image of $D$. Then, an AL strategy is used to assign a score $s^{AL}_i$ each image of $I_i$ from $D \smallsetminus D_l$.
    \item Given an annotation budget $b$, top-$b$ ranked images, w.r.t. $s^{AL}_i$, are returned to the user for annotation. The selected images are annotated by the user with regard to their relevance to the class-of-interest (1) or not (0).
    \item The newly labeled samples are added to $D_l$. Steps 4-7 are repeated until a stopping criterion.
\end{enumerate}

\subsection{Multi-object Setting}

Given a class $c$, a positive image is an image that contains at least one instance of $c$. A negative image doesn't contain any instance of $c$. A positive image $I_i$ could contain multiple objects, where each object is contained within its bounding box. In object-centric settings where the object dominates the whole image, a simple 1 or 0 label is sufficient. However, in our multi-object setting, we choose to query the user on the bounding box of the relevant object when present. For now, we assume that the user will select the most salient instance, i.e. the one with the largest bounding box. We denote this object as $o_i$. \\
\\
\indent The considered representation for the object $o_i$ should aim to characterize this object within a complex scene, without taking into account other possibly present objects, as the goal is to efficiently train a classifier that recognizes the desired pattern. The global representation $g_i$ is not fit for this goal, as it misses finer details and contains contextual information biased by surrounding objects, hence the need for a more details-focused representation. \\
\\
\indent In this work, we mainly focus on local patch-level descriptors. Each image $I_i$ is cropped into $M = 4, 16, ...$ regular patches $\{ I_i^1, I_i^2, ..., I_i^M \}$. To obtain the local representations $\{ l_i^1, l_i^2, ..., l_i^M \}$, we use 2 different methods:
\begin{itemize}
    \item We crop, resize, then feed each patch to the network: 
    \begin{equation}
        l_i^m = \Phi(I_i^m) \quad \forall m \in \{1, 2, ..., M\}
    \end{equation}
    \item We average the vision transformer patch tokens that belong to the patch.
\end{itemize}
We also compute an average pooled descriptor $\bar{l}_i$ from the local representations: 
\vspace{-0.5em}
\begin{equation}
    \bar{l}_i = avg_{m \in \{1, 2, ..., M\}} l_i^m
\end{equation}
The object $o_i$ can overlap with multiple patches from $\{ I_i^1, I_i^2, ..., I_i^M \}$. Let \\$\{1^+, 2^+, ..., M^+\}$ be the indices of the patches that overlap with the object $o_i$, and $\{1^-, 2^-, ..., M^-\}$ be the indices of the patches that don't. We have $M = M^+ + M^-$. Let $M^*$ be the index of the patch with the \textit{largest overlap} with the object $o_i$. In the next subsection, we detail the different representation strategies we suggest.

\subsection{Representation Strategies}
\subsubsection{Global-Only (GO).} 
Both negative and positive images are described using their global descriptors $g_i$.
\subsubsection{Local-Only (LO).}
\begin{itemize}
\setlength{\itemsep}{0em}
    \item \textbf{Local-Only-One (One sample per image):} each image only provides one sample. For positive images, we consider the descriptor of \textit{largest overlap} patch $l_i^{M^*}$. For negative images, we consider two alternatives: 1) \textit{Random patch} $l_i^m$ where $m \sim U(\{1, 2, ..., M\})$, or 2) \textit{Prototype of all patches} $\bar{l}_i$.
    \item \textbf{Local-Only-All (All samples per image):} each image results in many training samples. Negative images provide $M$ training samples $\{ l_i^1, l_i^2, ..., l_i^M \}$ all labeled 0. Positive images result in both positive AND negative samples: $\{ l_i^{1^+}, l_i^{2^+}, ...,l_i^{M^+} \}$ labeled 1 and $\{ l_i^{1^-}, l_i^{2^-}, ...,l_i^{M^-} \}$ labeled 0.
\end{itemize}
\subsubsection{Global-Local (GL).}
While previous strategies only relied on global or local contexts, global-local representation leverage both contexts to create one descriptor that combines both. We use two combination strategies: concatenation and pooling. Let $l_i^m$ be the chosen local descriptor, depending on the used local strategy. Concatenating the global descriptor $g_i$ with the local one results in a $2d$-dimensional vector $[g_i, l_i^m]$, but preserves both contexts separate. Average pooling $\frac{g_i + l_i^m}{2}$ keeps the same dimension, but merges the contexts. We use both combination strategies with each of the previous Local-Only representations. \\
\indent For the Local-Only-All representation, in the positive image case, concatenating the $g_i$ with each vector of $\{ l_i^{1^+}, l_i^{2^+}, ...,l_i^{M^+} \}$ labeled 1, then with each vector of $\{ l_i^{1^-}, l_i^{2^-}, ...,l_i^{M^-} \}$ labeled 0, results in similar descriptor parts-the global part-with different labels. In other words, as we are using an SVM classifier, if we concatenate global descriptors with positive patches and negatives patches, we observe that the shared global component between both positive and negative samples, during the training phase, introduces significant confusion in the classifier's weights estimation process, resulting in 0-weights for the global part. Therefore, to avoid this behavior, we only keep the positive part and concatenate $g_i$ with each vector of $\{ l_i^{1^+}, l_i^{2^+}, ...,l_i^{M^+} \}$.

\subsection{Inference and Active Learning Scores Computation}
After the classifier $f$ is trained on the annotated data, be it using global, local, or hybrid representations, $f$ is used to compute prediction scores of the dataset images. The idea is to keep one score $s_i$ per image $I_i \in D$. For the Global-Only representation, the prediction score is directly $s_i = f(g_i)$. However, for the Local-Only and Global-Local cases, as each image provides $M$ patches, the inference should be conducted on all patches. Each patch $m \in \{1, 2, ..., M\}$ results in a score $s_{i, m}$. We define the image score $s_i$ as: \vspace{-0.5em}
\begin{equation}
    s_i = \max_{m \in \{1, 2, ..., M\}} s_{i, m}
\end{equation}
Intuitively, if the object is present in the image, we expect the patches where the object is present to have high scores and the rest to have low scores, as a result the maximum score is what accounts for the object’s presence. If the maximum score is low, none of the patches contain the object. Note that when using prototypes for negatives, we also infer on the patches prototype $\bar{l}_i$, and include it in the computation of the maximum score. \\
\\
\indent To compute the AL score $s^{AL}_i$, we filter the labeled images, and only compute on $D \smallsetminus D_l$. In our work, we consider the uncertainty criterion~\cite{settles2009active}, as it is one of the most used and easiest to implement AL criteria. Also, our goal is rather the comparison of different representations. The AL score is given by: \vspace{-0.5em}
\begin{equation}
    s^{AL}_i = 1 - |0.5 - s_i| \quad \forall I_i \in D \smallsetminus D_l
\end{equation}

\subsection{Evaluation}
The aim of our task is to effectively retrieve relevant images containing the object of the class-of-interest. We are both interested in the retrieval performance of our framework, but also in the user satisfaction with the selected images. Hence, our evaluation is two-fold:
\begin{itemize}
\setlength{\itemsep}{0em}
    \item \textbf{The retrieval performance of the classifier:} we check if our classifier, using minimal data, is able to correctly retrieve information from a large dataset. We use the \textit{Mean Average Precision MAP} score on the inference scores $s_i$. 
    \item \textbf{The diversity of the selected positive images:} during the feedback phase, the user is queried on both positive and negative images. While these images should be informative to train the classifier, they should also satisfy the user. That is to say, positive images should be diverse and not exact copies of the query. We introduce the \textit{coverage} metric~\cite{zaher2026positive} to measure this diversity: we cluster the class-of-interest using a $k$-means clustering algorithm, then compute the proportion of clusters that contain at least one returned positive image. We use $k=32$, and average on $10$ clusterings for stable results. The coverage score ranges between 0 and 1, where a high coverage describes a set that covers many feature space regions of our class-of-interest.
\end{itemize}

\section{Experimental Results}

\subsection{Experimental Framework}

\subsubsection{Datasets and Features.} 
We conduct our experiments on two multi-object datasets: \textit{PascalVOC2012}~\cite{pascal-voc-2012} with $20$ object classes, and \textit{Coco2017}~\cite{lin2014microsoft} with $80$ object classes. We use the unsupervised \textit{dinoV2}~\cite{oquab2023dinov2} ViT, due to its remarkable retrieval performance, to extract both global and local features. Note that our focus on learning-free, interactive, multi-object, data-agnostic retrieval justifies comparing representations within our framework rather than against traditional retrieval baselines. As mentioned before, these baselines are data-specific as they are trained on a certain dataset, for a certain labeling.

\subsubsection{Query Preparation.} 
As the used datasets have predefined object classes, we query from each object class. For each class $c$, we create $Q=10$ queries, to ensure initial query diversity and stable results. A query is generated by randomly sampling $N_p=1$ positive image that contains an object of the class $c$, and $N_n=5$ negative images that don't. 

\subsubsection{Active Learning Loop.} 
For each query, we run the AL loop for $T=25$ iterations. We use a restricted annotation budget of $b=10$ images to return, in order to limit the annotation burden on the user. The user input is simulated using ground truth labels. As mentioned before, we use the uncertainty criterion, with a linear SVM classifier.

\subsection{Results}

In this subsection, we present the results obtained using the different representations. We evaluate the performance across the $T=25$ iterations by computing the area under the performance-iteration curve (AUC), obtained by trapezoidal integration and normalized by the number of iterations. 

\vspace{-1em}
\begin{table}[h]
    \centering
    \caption{Local-Only (LO) representations performances.}
    \label{global vs local2x2}
\begin{tabular}{|c|c||c|c|c||c|c|c|}
\hline
\multicolumn{2}{|c|}{\multirow{2}{*}{\makecell{Representation\\with $M=4 (2\times2)$}}} 
    & \multicolumn{3}{c||}{PascalVOC2012} 
    & \multicolumn{3}{c|}{Coco2017} \\
\cline{3-8}
\multicolumn{2}{|c|}{} 
    & MAP & Coverage & sum 
    & MAP & Coverage & sum \\
\hline \hline
\multicolumn{2}{|c||}{Global-Only} 
    & 0.829 & 0.652 & - 
    & 0.628 & 0.522 & - \\
\hline \hline

\multirow{3}{*}{\makecell{Feed\\forwarded\\crops}} 
    & LO-One-Proto 
        & \makecell{0.73\\(-11.94\%)} 
        & \makecell{0.142\\(-78.22\%)} 
        & -90.16\% 
        & \makecell{0.489\\(-22.13\%)} 
        & \makecell{0.095\\(-81.8\%)} 
        & -103.9\% \\
\cline{2-8}
& LO-One-Rand 
        & \makecell{0.866\\(4.46\%)} 
        & \makecell{0.618\\(-5.21\%)} 
        & -0.75\% 
        & \makecell{\textbf{0.683}\\\textbf{(8.76\%)}} 
        & \makecell{0.53\\(1.53\%)} 
        & 10.29\% \\
\cline{2-8}
& LO-All 
        & \makecell{0.865\\(4.34\%)} 
        & \makecell{0.634\\(-2.76\%)} 
        & 1.58\% 
        & \makecell{0.651\\(3.66\%)} 
        & \makecell{0.574\\(9.96\%)} 
        & 13.62\% \\
\hline \hline

\multirow{3}{*}{\makecell{Averaged\\patch\\tokens}} 
    & LO-One-Proto 
        & \makecell{0.668\\(-19.42\%)} 
        & \makecell{0.143\\(-78.07\%)} 
        & -97.49\% 
        & \makecell{0.403\\(-35.83\%)} 
        & \makecell{0.093\\(-82.18\%)} 
        & -118\% \\
\cline{2-8}
& LO-One-Rand 
        & \makecell{0.871\\(5.07\%)} 
        & \makecell{0.597\\(-8.44\%)} 
        & -3.37\% 
        & \makecell{0.663\\(5.57\%)} 
        & \makecell{0.514\\(-1.53\%)} 
        & 4.04\% \\
\cline{2-8}
& LO-All 
        & \makecell{\textbf{0.899}\\\textbf{(8.44\%)}} 
        & \makecell{\textbf{0.642}\\\textbf{(-1.53\%)}} 
        & \textbf{6.91\%} 
        & \makecell{0.676\\(7.64\%)} 
        & \makecell{\textbf{0.593}\\\textbf{(13.6\%)}} 
        & \textbf{21.24\%} \\
\hline
\end{tabular}
\end{table}

\vspace{-3em}
\subsubsection{Local Representation Selection.}
We present the results comparing the LO representations in Tab.~\ref{global vs local2x2}. We use $M=4$ patches to ensure fast learning and inference. Overall, the results indicate that the effectiveness of local representations strongly depends on how local information is encoded and aggregated. For both local feature extraction methods, the \textit{LO-One-Rand} and \textit{LO-All} strategies achieve competitive retrieval performance, consistently outperforming the global-only baseline in terms of MAP on both datasets. The \textit{LO-All} strategy yields clear improvements in coverage on \textit{Coco2017}, while these gains remain limited for \textit{PascalVOC2012}, where coverage is slightly lower than the global baseline. In contrast, the \textit{LO-One-Proto} variant performs poorly across all metrics, indicating that relying on a single prototype-based local descriptor is insufficient to capture the variability of object appearances in multi-object scenes. 

We also observe that averaging patch tokens generally leads to stronger and more stable performance compared to feed-forwarded crops. This improvement suggests that aggregating multiple patch-level descriptors provides a more robust and discriminative object representation for the active learning loop. Moreover, this encoding method offers a more efficient inference pipeline, as patch-level features can be extracted from a single forward pass of the ViT.

Overall, the averaged patch token with the \textit{LO-All} representation provides the best compromise between retrieval performance and coverage. We hypothesize that this is attributed to the larger and more diverse set of training samples available to the classifier during the AL process. Consequently, we adopt this strategy for the remainder of our experiments.

\vspace{-1em}
\begin{table}[h]
    \centering
    \caption{Global (GO) vs. Local (LO) vs. hybrid (GL) representations performances.}
\begin{tabular}{|c|c||c|c|c||c|c|c|}
\hline
\multicolumn{2}{|c|}{\multirow{2}{*}{\makecell{Representation\\with avg patch tokens}}} 
    & \multicolumn{3}{c||}{PascalVOC2012} 
    & \multicolumn{3}{c|}{Coco2017} \\
\cline{3-8}
\multicolumn{2}{|c|}{} 
    & MAP & Coverage & sum 
    & MAP & Coverage & sum \\
\hline \hline
\multicolumn{2}{|c||}{Global} 
    & 0.829 & 0.652 & - 
    & 0.628 & 0.522 & - \\
\hline \hline

\multirow{3}{*}{\makecell{2×2\\(4 patches)}} 
    & LO-All 
        & \makecell{0.899\\(8.44\%)} 
        & \makecell{0.642\\(-1.53\%)} 
        & 6.91\% 
        & \makecell{0.676\\(7.64\%)} 
        & \makecell{0.593\\(13.6\%)} 
        & 21.24\% \\
\cline{2-8}
& GL-concat-All 
        & \makecell{0.9\\(8.56\%)} 
        & \makecell{0.674\\(3.37\%)} 
        & 11.93\% 
        & \makecell{0.714\\(13.69\%)} 
        & \makecell{0.566\\(8.43\%)} 
        & 22.12\% \\
\cline{2-8}
& GL-pool-All 
        & \makecell{0.889\\(7.24\%)} 
        & \makecell{\textbf{0.695}\\\textbf{(6.6\%)}} 
        & \textbf{13.84\%} 
        & \makecell{0.674\\(7.32\%)} 
        & \makecell{0.596\\(14.18\%)} 
        & 21.5\% \\
\hline \hline

\multirow{3}{*}{\makecell{4×4\\(16 patches)}} 
    & LO-All 
        & \makecell{\textbf{0.913}\\\textbf{(10.13\%)}} 
        & \makecell{0.559\\(-14.26\%)} 
        & -4.13\% 
        & \makecell{0.711\\(13.22\%)} 
        & \makecell{0.576\\(10.34\%)} 
        & 23.56\% \\
\cline{2-8}
& GL-concat-All 
        & \makecell{0.907\\(9.41\%)} 
        & \makecell{0.652\\(0.0\%)} 
        & 9.41\% 
        & \makecell{\textbf{0.732}\\\textbf{(16.56\%)}} 
        & \makecell{0.564\\(8.05\%)} 
        & 24.61\% \\
\cline{2-8}
& GL-pool-All 
        & \makecell{0.896\\(8.08\%)} 
        & \makecell{0.651\\(-0.15\%)} 
        & 7.93\% 
        & \makecell{0.69\\(9.87\%)} 
        & \makecell{\textbf{0.603}\\\textbf{(15.52\%)}} 
        & \textbf{25.39\%} \\
\hline
\end{tabular}
    \label{global vs local vs global-local}
\end{table}

\vspace{-3em}
\subsubsection{Global-Only vs Local-Only vs Global-Local Representations.}
We check the improvements of using local or fusing global and local descriptors using all patches, w.r.t. to the global descriptor, in Tab.~\ref{global vs local vs global-local}, using $M=4$ and $M=16$ patches. Indeed, the fusion of descriptors results in better performances than using global or local representations alone, as showcased through Fig.~\ref{fig:example}. 

Across both datasets, hybrid representations consistently outperform local-only representations in terms of trade-off between both metrics and overall improvement, demonstrating the benefits of fusing global and local features, especially on \textit{PascalVOC2012}. These results highlight the complementary nature of global and local descriptors: concatenation favors retrieval precision, whereas pooling favors object coverage.

For \textit{PascalVOC2012}, using $M=4$ patches provides the best overall performance, balancing retrieval accuracy and coverage. In contrast, for \textit{COCO2017}, $M=16$ patches achieve higher MAP and coverage, indicating that a finer patch granularity is more beneficial for this larger and more diverse dataset. A more thorough discussion is provided later.

Overall, these results confirm that hybrid global-local representations are more effective than either global-only or local-only descriptors, and that the optimal patch size depends on the dataset characteristics. Furthermore, the choice of fusion strategy can be guided by the desired behavior: concatenation is preferable when high retrieval precision is prioritized, while pooling is advantageous when diverse object coverage is desired.

\subsubsection{The Effect of the Number of Patches.}
We analyze the effect of the number of patches based on the performance reported in Tab.~\ref{global vs local vs global-local}. For \textit{COCO2017}, using 16 patches improves both retrieval and coverage metrics. In contrast, for \textit{PascalVOC2012}, coverage is lower when using 16 patches compared to 4 patches or the global descriptor, despite similar retrieval performance. We hypothesize that this behavior is largely due to the object size distribution in each dataset.

Generally, in multi-objects datasets, objects are categorized into small, medium and large objects w.r.t their size within the image. We threshold the \{small, medium, large\} size categories using the number of patches, i.e. we use as thresholds $1/4 = 25\%$ and $1/16 = 6.25\%$. Tab.~\ref{object size dist} presents the percentage of each size category for both datasets. The table clearly shows the high proportion of small objects in \textit{Coco2017}, which is why the OR process greatly benefits from smaller patches. \textit{PascalVOC2012} on the other hand, has a more balanced size distribution, and using both 4 and 16 patches is beneficial. 

One possible solution is to concatenate the global descriptor with both the 4-patch and 16-patch local descriptors, providing flexibility to handle objects of varying sizes. However, this comes at the cost of a significantly slower interactive loop and higher memory usage. Another approach is to estimate object size distributions in a dataset using pre-trained segmentation models, which can inform the choice of patch granularity and help select an appropriate configuration before running the active learning loop.
\vspace{-1em}
\begin{table}[h]
    \centering
    \caption{Proportions of object size classes in each dataset.}
    \begin{tabular}{|c|c|c|c|}
        \hline
        Dataset & small (\footnotesize $<6.25\%$) & medium (\footnotesize $6.25-25\%$) & large (\footnotesize $>25\%$)\\
        \hline \hline
        Coco2017 & 74.7\% & 15.9\% & 9.4\% \\
        PascalVOC2012 & 44\% & 27.3\% & 28.7\% \\
        \hline
    \end{tabular}
    \vspace{0.3em}
    \label{object size dist}
\end{table}

\vspace{-3em}
\subsubsection{Iterative Analysis.}
\begin{figure}[h]
    \centering
    \includegraphics[width=0.98\linewidth]{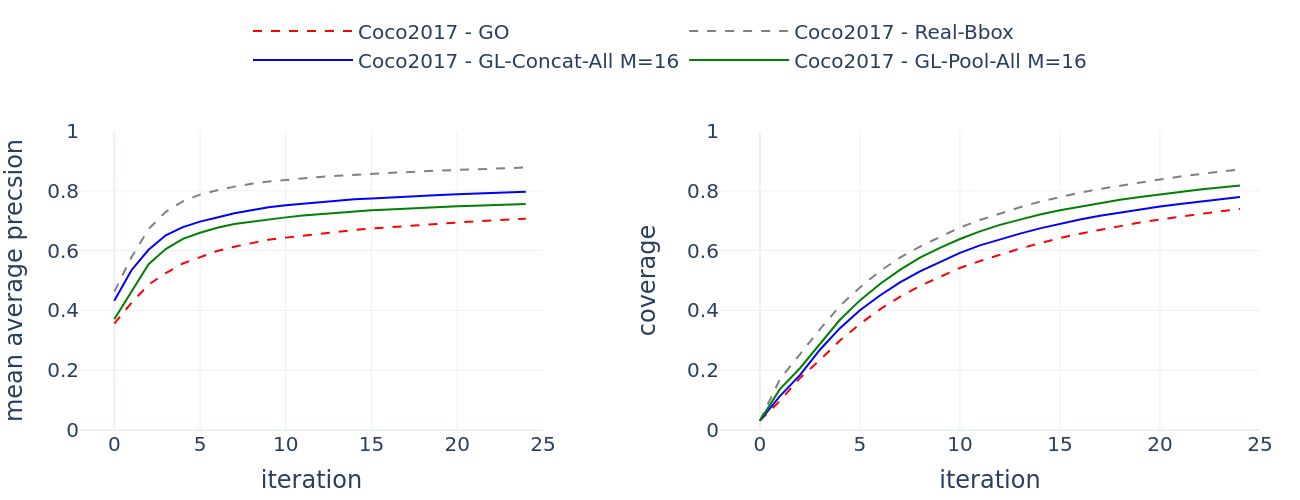}
    \caption{\centering Iterative performance on \textit{Coco2017}.}
    \label{fig:coco-across-iterations}
\end{figure}

\begin{figure}[h]
    \centering
    \includegraphics[width=0.98\linewidth]{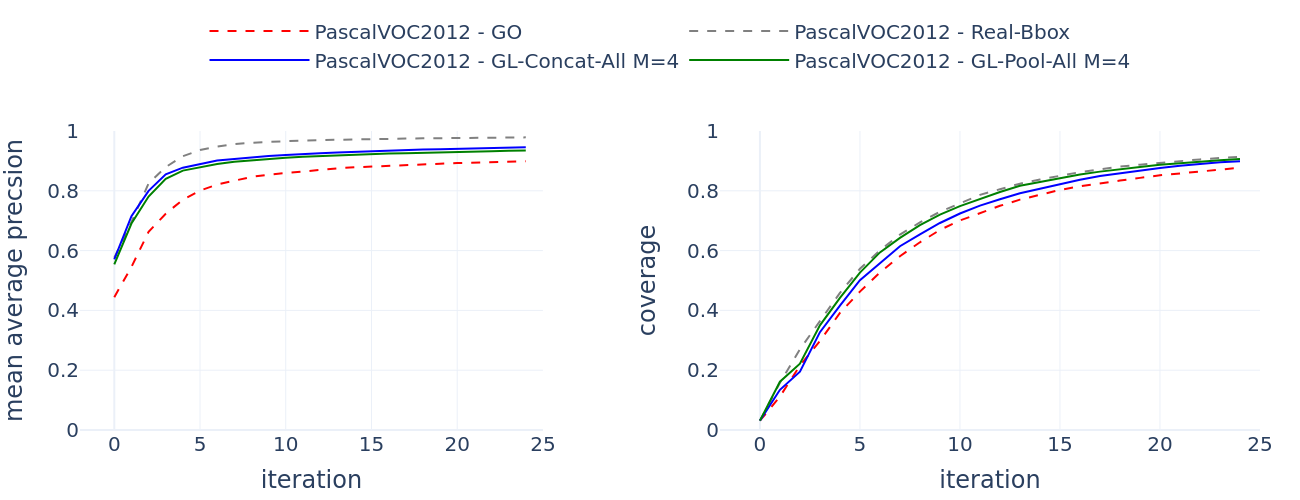}
    \caption{\centering Iterative performance on \textit{PascalVOC2012}.}
    \label{fig:pascal-across-iterations}
\end{figure}

We present the temporal progression of our metrics across iterations, for the best performance representations, in Fig.~\ref{fig:coco-across-iterations} and Fig.~\ref{fig:pascal-across-iterations}. We also report the performances using actual ground truth bounding boxes, which we do not have in a real life setting. These ground truth bounding boxes represent an upper bound for the expected performances. We observe that, for a target performance achieved by the global descriptor at $T=25$, our selected representations achieve this target since earlier iterations. This is particularly beneficial when the user wants to stop the iterative process earlier. Moreover, our selected representations showcases significant retrieval gains, compared to the global descriptor across both datasets.
\vspace{-2em}

\subsubsection{The case of Vision Transformers with Registers.} Register-based ViTs~\cite{darcet2023vision}, as an extension of standard ViTs, have been shown to improve dense prediction tasks such as Object Discovery. As our setup share some characteristics with dense tasks, we investigate the use of such models. We use the same architecture and unsupervised training framework to compare. We report the score in Tab.~\ref{tab: reg global vs local vs global-local}. Overall, comparing to the results of Tab.~\ref{global vs local vs global-local}, standard ViTs show higher performance. We hypothesize that entangled global information in local patch tokens benefits the performance, especially helping to ensure better diversity. For ViTs with registers, fusing global and local features still improves results for \textit{PascalVOC2012} where objects are fewer and less crowded. However, the gains are smaller on \textit{Coco2017}, where complex scenes reduce the need for a general global context of the whole image. Due to their multiple attention links to register tokens, patch tokens now carry more structured nearby global information, explicitely aggregated and conditioned via the registers, which may explain why \textit{LO-ALL} remains competitive.
\vspace{-1em}
\begin{table}[h]
\label{tab: reg global vs local vs global-local}
    \centering
    \caption{Global (GO) vs. Local (LO) vs. hybrid (GL) representations performances.}
\begin{tabular}{|c||c|c|c||c|c|c|}
\hline
\multirow{2}{*}{\makecell{Representation\\with avg patch tokens}} 
    & \multicolumn{3}{c||}{PascalVOC2012 (2$\times$2)} 
    & \multicolumn{3}{c|}{COCO2017 (4$\times$4)} \\
\cline{2-7}
    & MAP & Coverage & sum 
    & MAP & Coverage & sum \\
\hline \hline

Global 
    & 0.823 & 0.626 & - 
    & 0.626 & 0.506 & - \\
\hline \hline

LO-All 
    & \makecell{0.900\\(9.36\%)} 
    & \makecell{0.602\\(-3.83\%)} 
    & 5.53\% 
    & \makecell{0.730\\(16.61\%)} 
    & \makecell{0.575\\(13.64\%)} 
    & \textbf{30.25\%} \\
\hline

GL-concat-All 
    & \makecell{0.900\\(9.36\%)} 
    & \makecell{0.635\\(1.44\%)} 
    & 10.80\% 
    & \makecell{\textbf{0.739}\\\textbf{(18.05\%)}} 
    & \makecell{0.550\\(8.70\%)} 
    & 26.75\% \\
\hline

GL-pool-All 
    & \makecell{0.885\\(7.53\%)} 
    & \makecell{\textbf{0.659}\\\textbf{(5.27\%)}} 
    & \textbf{12.80\%} 
    & \makecell{0.699\\(11.66\%)} 
    & \makecell{\textbf{0.599}\\\textbf{(18.38\%)}} 
    & 30.04\% \\
\hline
\end{tabular}
\end{table}
\vspace{-2em}

\section{Conclusion and Future work}

In this work, we presented the Human-in-the-Loop Object Retrieval framework, an interactive approach to retrieving objects from large, unlabeled datasets, in the multi-object image setting. By leveraging Active Learning within an interactive feedback loop, our approach efficiently balances retrieval performance and user effort. We demonstrated the efficiency of our framework, and studied different representations strategies for our images. Our finding showcased the importance of including global and local descriptors through concatenation, in order to allow our classier to separately consider both the global and local contexts. In addition, we showed the potential of such representations for practical applications where user time is critical to accelerate the retrieval process of objects. Further works include the use of hierarchic local descriptors to adapt to different objects sizes. We would like to design an adaptive multi-scale patch Active Selection framework that dynamically identifies salient regions across varying patch resolutions to optimize retrieval efficiency in cluttered multi-object scenes.
%
%
%
%
\bibliographystyle{splncs04}
\bibliography{main}

@inproceedings{arandjelovic2016netvlad,
  title={NetVLAD: CNN architecture for weakly supervised place recognition},
  author={Arandjelovic, Relja and Gronat, Petr and Torii, Akihiko and Pajdla, Tomas and Sivic, Josef},
  booktitle={Proceedings of the IEEE conference on computer vision and pattern recognition},
  pages={5297--5307},
  year={2016}
}

@inproceedings{deselaers2008overview,
  title={Overview of the ImageCLEF 2007 object retrieval task},
  author={Deselaers, Thomas and Hanbury, Allan and Viitaniemi, Ville and Bencz{\'u}r, Andr{\'a}s and Brendel, M{\'a}ty{\'a}s and Dar{\'o}czy, B{\'a}lint and Escalante Balderas, Hugo Jair and Gevers, Theo and Hern{\'a}ndez Gracidas, Carlos Arturo and Hoi, Steven CH and others},
  booktitle={Advances in Multilingual and Multimodal Information Retrieval: 8th Workshop of the Cross-Language Evaluation Forum, CLEF 2007, Budapest, Hungary, September 19-21, 2007, Revised Selected Papers 8},
  pages={445--471},
  year={2008},
  organization={Springer}
}

@inproceedings{litayem2009interactive,
  title={Interactive objects retrieval with efficient boosting},
  author={Litayem, Saloua and Joly, Alexis and Boujemaa, Nozha},
  booktitle={Proceedings of the 17th ACM international conference on Multimedia},
  pages={545--548},
  year={2009}
}

@article{tolias2015particular,
  title={Particular object retrieval with integral max-pooling of CNN activations},
  author={Tolias, Giorgos and Sicre, Ronan and J{\'e}gou, Herv{\'e}},
  journal={arXiv preprint arXiv:1511.05879},
  year={2015}
}

@inproceedings{arandjelovic2012three,
  title={Three things everyone should know to improve object retrieval},
  author={Arandjelovi{\'c}, Relja and Zisserman, Andrew},
  booktitle={2012 IEEE conference on computer vision and pattern recognition},
  pages={2911--2918},
  year={2012},
  organization={IEEE}
}

@article{lowe2004distinctive,
  title={Distinctive image features from scale-invariant keypoints},
  author={Lowe, David G},
  journal={International journal of computer vision},
  volume={60},
  pages={91--110},
  year={2004},
  publisher={Springer}
}

@inproceedings{csurka2004visual,
  title={Visual categorization with bags of keypoints},
  author={Csurka, Gabriella and Dance, Christopher and Fan, Lixin and Willamowski, Jutta and Bray, C{\'e}dric},
  booktitle={Workshop on statistical learning in computer vision, ECCV},
  volume={1},
  number={1-22},
  pages={1--2},
  year={2004},
  organization={Prague}
}

@inproceedings{perronnin2007fisher,
  title={Fisher kernels on visual vocabularies for image categorization},
  author={Perronnin, Florent and Dance, Christopher},
  booktitle={2007 IEEE conference on computer vision and pattern recognition},
  pages={1--8},
  year={2007},
  organization={IEEE}
}

@inproceedings{jegou2010aggregating,
  title={Aggregating local descriptors into a compact image representation},
  author={J{\'e}gou, Herv{\'e} and Douze, Matthijs and Schmid, Cordelia and P{\'e}rez, Patrick},
  booktitle={2010 IEEE computer society conference on computer vision and pattern recognition},
  pages={3304--3311},
  year={2010},
  organization={IEEE}
}

@inproceedings{razavian2014cnn,
  title={CNN features off-the-shelf: an astounding baseline for recognition},
  author={Razavian, Ali Sharif and Azizpour, Hossein and Sullivan, Josephine and Carlsson, Stefan},
  booktitle={Proceedings of the IEEE conference on computer vision and pattern recognition workshops},
  pages={806--813},
  year={2014}
}

@article{el2021training,
  title={Training vision transformers for image retrieval},
  author={El-Nouby, Alaaeldin and Neverova, Natalia and Laptev, Ivan and J{\'e}gou, Herv{\'e}},
  journal={arXiv preprint arXiv:2102.05644},
  year={2021}
}

@inproceedings{song2023boosting,
  title={Boosting vision transformers for image retrieval},
  author={Song, Chull Hwan and Yoon, Jooyoung and Choi, Shunghyun and Avrithis, Yannis},
  booktitle={Proceedings of the IEEE/CVF winter conference on applications of computer vision},
  pages={107--117},
  year={2023}
}

@article{chen2022deep,
  title={Deep learning for instance retrieval: A survey},
  author={Chen, Wei and Liu, Yu and Wang, Weiping and Bakker, Erwin M and Georgiou, Theodoros and Fieguth, Paul and Liu, Li and Lew, Michael S},
  journal={IEEE Transactions on Pattern Analysis and Machine Intelligence},
  volume={45},
  number={6},
  pages={7270--7292},
  year={2022},
  publisher={IEEE}
}

@inproceedings{tan2021instance,
  title={Instance-level image retrieval using reranking transformers},
  author={Tan, Fuwen and Yuan, Jiangbo and Ordonez, Vicente},
  booktitle={proceedings of the IEEE/CVF international conference on computer vision},
  pages={12105--12115},
  year={2021}
}

@article{patil2011relevance,
  title={Relevance Feedback in Content Based Image Retrieval: A Review.},
  author={Patil, Pushpa B and Kokare, Manesh B},
  journal={Journal of Applied Computer Science \& Mathematics},
  number={10},
  year={2011}
}

@article{ngo2016image,
  title={Image retrieval with relevance feedback using SVM active learning},
  author={Ngo, Giang Truong and Ngo, Tao Quoc and Nguyen, Dung Duc},
  journal={International Journal of Electrical and Computer Engineering},
  volume={6},
  number={6},
  pages={3238},
  year={2016},
  publisher={IAES Institute of Advanced Engineering and Science}
}

@inproceedings{tong2001support,
  title={Support vector machine active learning for image retrieval},
  author={Tong, Simon and Chang, Edward},
  booktitle={Proceedings of the ninth ACM international conference on Multimedia},
  pages={107--118},
  year={2001}
}

@article{settles2009active,
  title={Active learning literature survey},
  author={Settles, Burr},
  year={2009},
  publisher={University of Wisconsin-Madison Department of Computer Sciences}
}

@article{ferecatu2007interactive,
  title={Interactive remote-sensing image retrieval using active relevance feedback},
  author={Ferecatu, Marin and Boujemaa, Nozha},
  journal={IEEE Transactions on Geoscience and Remote Sensing},
  volume={45},
  number={4},
  pages={818--826},
  year={2007},
  publisher={IEEE}
}

@article{demir2014novel,
  title={A novel active learning method in relevance feedback for content-based remote sensing image retrieval},
  author={Demir, Beg{\"u}m and Bruzzone, Lorenzo},
  journal={IEEE Transactions on Geoscience and Remote Sensing},
  volume={53},
  number={5},
  pages={2323--2334},
  year={2014},
  publisher={IEEE}
}

@article{aggarwal2022optimizing,
  title={Optimizing active learning for low annotation budgets},
  author={Aggarwal, Umang and Popescu, Adrian and Hudelot, C{\'e}line},
  journal={arXiv preprint arXiv:2201.07200},
  year={2022}
}

@inproceedings{abdali2023active,
  title={Active learning for efficient few-shot classification},
  author={Abdali, Aymane and Gripon, Vincent and Drumetz, Lucas and Boguslawski, Bartosz},
  booktitle={ICASSP 2023-2023 IEEE International Conference on Acoustics, Speech and Signal Processing (ICASSP)},
  pages={1--5},
  year={2023},
  organization={IEEE}
}

@article{wang2020generalizing,
  title={Generalizing from a few examples: A survey on few-shot learning},
  author={Wang, Yaqing and Yao, Quanming and Kwok, James T and Ni, Lionel M},
  journal={ACM computing surveys (csur)},
  volume={53},
  number={3},
  pages={1--34},
  year={2020},
  publisher={ACM New York, NY, USA}
}

@misc{pascal-voc-2012,
	author = "Everingham, M. and Van~Gool, L. and Williams, C. K. I. and Winn, J. and Zisserman, A.",
	title = "The {PASCAL} {V}isual {O}bject {C}lasses {C}hallenge 2012 {(VOC2012)} {R}esults",
	howpublished = "http://www.pascal-network.org/challenges/VOC/voc2012/workshop/index.html"}

@inproceedings{lin2014microsoft,
  title={Microsoft coco: Common objects in context},
  author={Lin, Tsung-Yi and Maire, Michael and Belongie, Serge and Hays, James and Perona, Pietro and Ramanan, Deva and Doll{\'a}r, Piotr and Zitnick, C Lawrence},
  booktitle={Computer vision--ECCV 2014: 13th European conference, zurich, Switzerland, September 6-12, 2014, proceedings, part v 13},
  pages={740--755},
  year={2014},
  organization={Springer}
}

@article{oquab2023dinov2,
  title={Dinov2: Learning robust visual features without supervision},
  author={Oquab, Maxime and Darcet, Timoth{\'e}e and Moutakanni, Th{\'e}o and Vo, Huy and Szafraniec, Marc and Khalidov, Vasil and Fernandez, Pierre and Haziza, Daniel and Massa, Francisco and El-Nouby, Alaaeldin and others},
  journal={arXiv preprint arXiv:2304.07193},
  year={2023}
}

@article{dosovitskiy2020image,
  title={An image is worth 16x16 words: Transformers for image recognition at scale},
  author={Dosovitskiy, Alexey and Beyer, Lucas and Kolesnikov, Alexander and Weissenborn, Dirk and Zhai, Xiaohua and Unterthiner, Thomas and Dehghani, Mostafa and Minderer, Matthias and Heigold, Georg and Gelly, Sylvain and others},
  journal={arXiv preprint arXiv:2010.11929},
  year={2020}
}

@article{o2015introduction,
  title={An introduction to convolutional neural networks},
  author={O'shea, Keiron and Nash, Ryan},
  journal={arXiv preprint arXiv:1511.08458},
  year={2015}
}

@inproceedings{garg2024revisit,
  title={Revisit Anything: Visual Place Recognition via Image Segment Retrieval},
  author={Garg, Kartik and Puligilla, Sai Shubodh and Kolathaya, Shishir and Krishna, Madhava and Garg, Sourav},
  booktitle={European Conference on Computer Vision},
  pages={326--343},
  year={2024},
  organization={Springer}
}

@article{mosqueira2023human,
  title={Human-in-the-loop machine learning: a state of the art},
  author={Mosqueira-Rey, Eduardo and Hern{\'a}ndez-Pereira, Elena and Alonso-R{\'\i}os, David and Bobes-Bascar{\'a}n, Jos{\'e} and Fern{\'a}ndez-Leal, {\'A}ngel},
  journal={Artificial Intelligence Review},
  volume={56},
  number={4},
  pages={3005--3054},
  year={2023},
  publisher={Springer}
}

@inproceedings{shao2023global,
  title={Global features are all you need for image retrieval and reranking},
  author={Shao, Shihao and Chen, Kaifeng and Karpur, Arjun and Cui, Qinghua and Araujo, Andr{\'e} and Cao, Bingyi},
  booktitle={Proceedings of the IEEE/CVF International Conference on Computer Vision},
  pages={11036--11046},
  year={2023}
}

@inproceedings{yang2021dolg,
  title={Dolg: Single-stage image retrieval with deep orthogonal fusion of local and global features},
  author={Yang, Min and He, Dongliang and Fan, Miao and Shi, Baorong and Xue, Xuetong and Li, Fu and Ding, Errui and Huang, Jizhou},
  booktitle={Proceedings of the IEEE/CVF International conference on Computer Vision},
  pages={11772--11781},
  year={2021}
}

@inproceedings{cao2020unifying,
  title={Unifying deep local and global features for image search},
  author={Cao, Bingyi and Araujo, Andre and Sim, Jack},
  booktitle={Computer Vision--ECCV 2020: 16th European Conference, Glasgow, UK, August 23--28, 2020, Proceedings, Part XX 16},
  pages={726--743},
  year={2020},
  organization={Springer}
}

@article{darcet2023vision,
  title={Vision transformers need registers},
  author={Darcet, Timoth{\'e}e and Oquab, Maxime and Mairal, Julien and Bojanowski, Piotr},
  journal={arXiv preprint arXiv:2309.16588},
  year={2023}
}

@article{song2022dalg,
  title={Dalg: Deep attentive local and global modeling for image retrieval},
  author={Song, Yuxin and Zhu, Ruolin and Yang, Min and He, Dongliang},
  journal={arXiv preprint arXiv:2207.00287},
  year={2022}
}

@inproceedings{govindarajan2021self,
  title={Self-Supervised Representation Learning for Content Based Image Retrieval of Complex Scenes},
  author={Govindarajan, Hariprasath and Lindskog, Peter and Lundstr{\"o}m, Dennis and Olmin, Amanda and Roll, Jacob and Lindsten, Fredrik},
  booktitle={2021 IEEE Intelligent Vehicles Symposium Workshops (IV Workshops)},
  pages={249--256},
  year={2021},
  organization={IEEE}
}

@article{denner2025leveraging,
  title={Leveraging foundation models for content-based image retrieval in radiology},
  author={Denner, Stefan and Zimmerer, David and Bounias, Dimitrios and Bujotzek, Markus and Xiao, Shuhan and Stock, Raphael and Kausch, Lisa and Schader, Philipp and Penzkofer, Tobias and J{\"a}ger, Paul F and others},
  journal={Computers in Biology and Medicine},
  volume={196},
  pages={110640},
  year={2025},
  publisher={Elsevier}
}

@inproceedings{zaher2026positive,
  title={Positive-First Most Ambiguous: A Simple Active Learning Criterion for Interactive Retrieval of Rare Categories},
  author={Zaher, Kawtar and Buisson, Olivier and Joly, Alexis},
  booktitle={CVPRW 2026 - The 13th Workshop on Fine-Grained Visual Categorization (FGVC13)},
  year={2026}
}
\end{document}